\documentclass{article}

\usepackage[preprint]{neurips_2026}
\workshoptitle{Sim2Science}

\usepackage[utf8]{inputenc}
\usepackage[T1]{fontenc}
\usepackage{microtype}
\usepackage{amsmath,amssymb}
\usepackage{textgreek}
\usepackage{booktabs}
\usepackage{float}
\usepackage{siunitx}
\usepackage{caption}
\usepackage{xcolor}
\usepackage{adjustbox}
\usepackage{listings}
\usepackage{graphicx}
\usepackage{tikz}
\usetikzlibrary{arrows.meta, shapes, positioning, calc}
\usepackage[english]{babel}
\usepackage{csquotes}
\usepackage{hyperref}
\usepackage{cleveref}
\usepackage{subcaption}
\usepackage{listings}

\definecolor{RowGray}{gray}{0.94}
\definecolor{instadeepblue}{HTML}{006eb8}
\definecolor{codegray}{gray}{0.95}

\title{ALF: An Active Learning Framework for Scientific Discovery}

\author{%
  Shikha Surana\thanks{Correspondence: \texttt{s.surana@instadeep.com}}\\
  InstaDeep\\
  \And
  Alex Hawkins-Hooker\thanks{Work completed during an internship at InstaDeep.}\\
  University College London\\
  \And
  Olivia Gallup\footnotemark[2]\\
  University of Oxford\\
  \And
  Christoph Brunken\\
  InstaDeep\\
  \And
  Jules Tilly\\
  InstaDeep\\
  \And
  Paul Duckworth\\
  InstaDeep\\
}

\begin{document}
\maketitle

\begin{abstract} \looseness=-1
Machine learning for scientific discovery is almost systematically data bound. Producing relevant high quality data, under budget constraints, is amongst the most promising ways to advance the field. Active learning (AL) offers promise wherever labelling requires expensive experiment, measurement, or simulation. Most existing tools cover only part of the data acquisition loop, and typically focus on either offline benchmarking or online deployment, but not both. We present \textsc{ALF}, a modular AL Framework that runs the full data acquisition loop via five modular components. One clear API for both settings: \textit{offline}, against an existing dataset for controlled and reproducible experimentation; and \textit{online}, against an oracle for acquiring new candidates in real-world deployments. ALF is open-source and available at \url{https://github.com/instadeepai/alf}.
\end{abstract}



\section{Introduction} \looseness=-1
\label{sec:introduction}

Across many scientific domains, large combinatorial search-spaces mean that exhaustive data screening is rarely possible. Coupled with potentially high cost of acquiring new measurements, e.g. wet-lab assays or MD simulations, the practical problem becomes one of targeted data acquisition. That is, given budget constraints, \textit{what new data should we acquire to maximally learn about our objective?}

\looseness=-1
Active learning (AL)~\citep{settles2011theories} aims to address this question through an iterative campaign. In each round, a \emph{surrogate} model is fit on the current (labelled) dataset. Next, an \emph{acquisition function} uses the model's posterior predictive distribution to score how informative future candidates (unlabelled data) are expected to be on the objective. An \emph{oracle} then labels those selected candidates, and they, along with their labels, are added to the training dataset for the next round. The oracle is the reference the campaign treats as ground truth. The goal of the campaign is to maximally utilise its constrained measurement budget to best serve the objective. 
Through careful treatment of acquisition strategies, the same data acquisition loop can support Bayesian optimisation (BO) objectives also~\citep{williams2006gaussian, garnett2023bayesian}, scoring unlabelled candidates under the posterior most likely to maximise an unknown function, rather than AL information gain strategies.

\looseness=-1
In practice, data acquisition is typically benchmarked \emph{offline} against an available labelled dataset, then deployed \emph{online} against an expensive oracle of interest. Yet, these two settings are usually served by different and incompatible tools. Offline, the possible search space is derived from a static (labelled) dataset, and search is reduced to evaluating over the dataset and the oracle is a simple lookup table. Online however, a search function or generative model is required to propose new candidates each round, and the oracle is queried to score those selected by the acquisition function. We posit that the same data acquisition loop can be utilised for both settings. We present ALF, an ask/tell framework for scientific discovery and make the following technical contributions: 
\vspace{-0.5em}
\begin{enumerate}
 \setlength{\itemsep}{0pt}\setlength{\parskip}{0pt}
    \item \textbf{Single offline and online API}.
    ALF provides a single, clean API for offline campaigns (against a dataset), as for online campaigns (against an oracle). This allows for detailed benchmarking and rapid real-world deployment. 

    \item \textbf{Modular architecture.} ALF provides five modular and interchangeable components: datasets, surrogates, acquisition functions, search functions, and oracle classes. Each with standard, lightweight interfaces, allowing for extensions and plug-and-play experimentation. 

    \item \textbf{Data type agnostic.} ALF is build invariant to data modality, which means the same clean API can be utilised across multiple scientific domains as we show in \cref{sec:use-cases}.
\end{enumerate}
\vspace{-0.5em}

\section{Related Work} \label{sec:related_work}

\looseness=-1
Active learning~\citep{settles2011theories} and Bayesian optimisation~\citep{williams2006gaussian, garnett2023bayesian} are popular methods for guided data acquisition, however current frameworks address a subset of the typical scientific discovery loop. 
Ax~\citep{Bakshy2018AEAD}, built on the BoTorch~\citep{balandat2020botorch} acquisition backend that ALF also uses, targets continuous parameter and hyperparameter optimisation. General AL libraries such as modAL~\citep{modAL2018} does provide model wrappers and queryable strategies for tabular data, however, neither Ax nor modAL ship surrounding scientific data acquisition tools such as data handling, labelling, and multi-round framework. ALF supplies these features as interchangeable components, utilising a single and clean API for both offline and online data acquisition tasks.
FLEXS~\citep{sinai2020adalead}, MolPAL~\citep{graff2021molpal} and ALDE~\citep{yang2025active} provide a sequential AL loop but are limited to biological sequences and small-molecule screening respectively, i.e. neither is domain-agnostic.

\looseness=-1
GAUCHE~\citep{griffiths2023gauche} contributes GP kernels over sequences, molecular graphs and strings; however its focus is surrogate design rather than sequential campaigns, and ~\cite{benjamins2024bayesian} demonstrates that such kernels can be seamlessly incorporated into ALF's surrogate class. 
Domain toolkits such as TorchDrug~\citep{zhu2022torchdrug} and DeepChem~\citep{Ramsundar-et-al-2019} pair curated datasets with model implementations and training engines, but do not provide acquisition functions or oracle abstractions to close the feedback loop. 
Closest to ALF, MADE~\citep{malik2026made} runs closed-loop discovery with composable components, but is specific to material domains and is built to evaluate discovery \textit{agents} under budget constraints. These libraries do not provide modular components spanning the full discovery loop that can be run unchanged across modalities and in both offline and online settings.
\section{ALF: Active Learning Framework}
\label{sec:framework}

\subsection{Framework} 
\begin{figure}[h!]
\centering
\includegraphics[width=\linewidth]{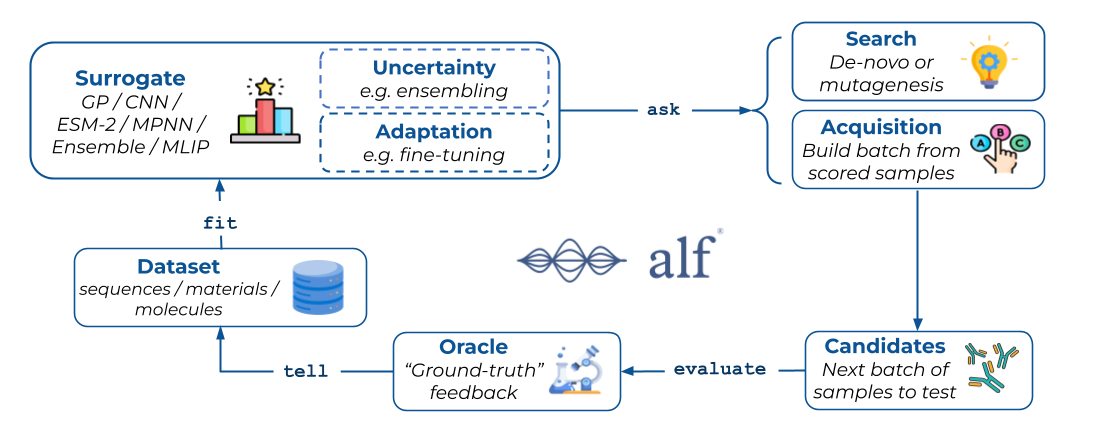}
\caption{\textbf{ALF: Data acquisition loop.} 
At each round, a surrogate model is first \textit{fit} then \textit{queried} (ask) to iteratively approximate an oracle of interest.} 
\label{fig:alf_loop}
\end{figure}

\looseness=-1
\textsc{ALF} runs a data acquisition campaign via a modular ask/tell framework, shown in \cref{fig:alf_loop}, where each component can be easily interchanged. 
A design campaign is orchestrated by the \texttt{DesignTask}, and comprises of a fixed number of rounds. 
Each round comprises of the following steps: (1) the surrogate is \textbf{fit} to the dataset collected so far, (2) an acquisition function  \textbf{scores} new candidates, (3) the oracle  \textbf{evaluates} the top-scoring candidates, and finally, (4) the train dataset is \textbf{augmented} with these newly labelled candidates. 
The \texttt{State} object is updated each round and manages the dataset (including train, validation, test, and candidate pool splits), the surrogate, round index, acquisition history, and all recorded metrics. Fig~\ref{fig:api} shows the minimal code to run a design campaign on GFP protein sequences.

\subsection{Components} 
\textsc{ALF} is shipped as two complementary packages: \texttt{alf-core} contains all the base classes, with ready-to-use example implementations in \texttt{alf-tools}.
We go into more detail of each component next, and Appendix~\ref{app:library} further describes the library, the package structure, the component base classes, task types, offline-to-online transition, extending \textsc{ALF}, and available tutorials and documentation.

\begin{itemize}

    \item \textbf{Dataset class} maintains and manages the candidate data. It implements train/validation/test/pool splits that the design loop draws from. It is data-type agnostic, and new datasets require only a lightweight data loader. We provide implementations in \texttt{alf-tools} for protein sequences: GFP~\cite{sarkisyan2016gfp}, ProteinGym~\cite{notin2023proteingym}, and FLIP~\cite{dallago2021flip}; small molecules: GuacaMol~\cite{brown2019guacamol}; and materials: MatBench~\cite{dunn2020matbench}.

    \item \textbf{Surrogate class} is responsible for featurising and modelling the currently available labelled train dataset via wrapping a \texttt{BaseModel}. Once fit, it is then used to predict target values for unlabelled test candidates. \texttt{alf-tools} includes example implementations of CNN, MLP, GP, ESM-2~\cite{lin2023esm2}, Chemprop MPNN~\cite{yang2019chemprop}, MLIP (\texttt{mlip} library~\citep{brunken2026machinelearninginteratomicpotentials} \& MACE~\cite{batatia2022mace}), and \texttt{EnsembleWrapper}.

    \item \textbf{Search function} is the class responsible for populating the candidate pool at each round. In the \textit{offline} setting, this function reduces to a simple \texttt{DatasetSearch}. However, in the \textit{online} setting, it is responsible for generating new unlabelled candidates with which to score and select from. We provide a framework for popular approaches to:
    1) \texttt{ProtocolSearch} augment known candidates via a protocol, and 2) \texttt{GeneratorSearch} utilise a generative model to propose new candidates directly. Both are supported through a single API. 
    In \texttt{alf-tools} we provide implementations for \texttt{SingleMutantSearch} (sequences), \texttt{SmilesMutationSearch} (molecules), and \texttt{ElementSubstitutionSearch} (materials). 

    \item \textbf{Acquisition function} scores each candidate in the candidate pool by its utility: an estimate, under the surrogate model, of how valuable it would be to acquire the ground truth value. Implementations available include Greedy, UCB~\cite{srinivas2010gpucb}, EI~\cite{jones1998ei}, Thompson sampling (TS)~\cite{russo2018thompson}, and CoreSet~\cite{sener2018coreset}, alongside a BoTorch backend for easy integration of new acquisition functions.

    \item \textbf{Oracle class} provides ground truth labels for each of the acquired candidates. In the \textit{offline} setting it wraps a \texttt{BaseDataset} object to return the relevant recorded values, and in the \textit{online} setting it wraps a \texttt{BaseModel} to compute the labels on the fly. Model implementations include ESMFold~\cite{lin2023esm2}, PyRosetta~\cite{chaudhury2010pyrosetta}, and an RDKit molecular property predictor~\cite{rdkit}.
\end{itemize}

\begin{figure}[h]
\begin{lstlisting}
config    = BaseDatasetConfig(name="gfp", modality="sequence", ...)
dataset   = GFP(config)
surrogate = Surrogate(model=ESM2Model())
optimizer = Optimizer(acquisition_fn=Greedy(), search_fn=DatasetSearch())
oracle    = Oracle(scorer=dataset)          # offline: dataset lookup
 
task  = DesignTask(num_acq_rounds=5, acq_batch_size=100)
state = task.setup(dataset=dataset, surrogate=surrogate)
task.run(state=state, optimizer=optimizer, oracle=oracle)
\end{lstlisting}
\caption{An offline, 5-round, \texttt{DesignTask} implemented in \textsc{ALF} utilising pre-trained ESM2 surrogate. To switch to \textit{online} mode, only the search and oracle lines need changing (see Appendix~\ref{app:offline_online}).}
\label{fig:api}
\end{figure}

\section{Case Studies} \label{sec:use-cases}

\subsection{Experimental Setup}
\label{subsec:setup}

\looseness=-1
In this section we demonstrate \textsc{ALF}'s current capabilities. We present three popular multi-round design tasks and demonstrate each as an \textit{offline} data acquisition benchmark, and then directly adapt them to an \textit{online} campaign for real-world deployment. We highlight the modality-agnostic flexibility, and present results across both active learning and Bayesian optimisation objectives. 

\textbf{Design Tasks}: 1) Protein sequence design: \texttt{HIS7\_YEAST} a multi-mutant protein-landscape taken from the ProteinGym benchmark~\citep{notin2023proteingym};  2) Small molecule design: GuacaMol \texttt{osimertinib\_mpo} objective \citep{brown2019guacamol}, and 3) Material design: \texttt{matbench\_mp\_e\_form} taken from MatBench~\citep{dunn2020matbench}. 

Each design campaign begins with an initial 5,000 labelled candidates as ``train'' dataset, runs for 10 acquisition rounds and acquires 25 new candidates per round. 
We use a Gaussian process (GP) surrogate model (\texttt{GPModel}), and only the input data featuriser varies with modality (details in \cref{app:exp_details}). 
For AL objectives, we highlight two entropy-based acquisition strategies (Uncertainty sampling~\citep{settles2011theories} and CoreSet~\citep{sener2018coreset}), and report mean RMSE at each round (lower is better). 
For BO objectives, four utility-based acquisition strategies (UCB, EI, TS and Greedy), and report the max score acquired so far at each round (higher is better). All experiments are repeated for 5 random seeds.

\subsection{Offline Benchmarking}

\begin{figure}[h!]
\centering
\includegraphics[width=0.8\linewidth]{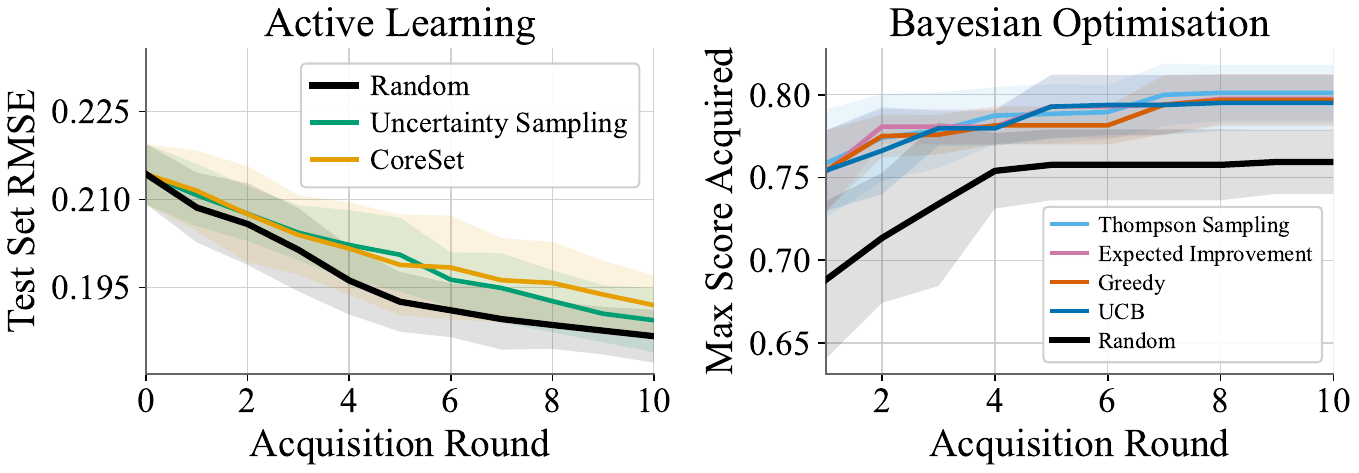}
\caption{\textbf{Small molecule design}. Offline benchmarking of Active learning objective (left) and Bayesian optimisation (right). Results are the mean over $5$ random seeds; error bars are $\pm 1$ std.}
\label{fig:offline}
\end{figure}

We highlight \textsc{ALF}'s offline benchmark capabilities across seven acquisition strategies for AL and BO objectives. 
Here, the offline dataset provides both the initial training dataset and the candidate pool with which to acquire new ground truth data from -- helping to facilitate clear and actionable insights across methodologies. 
\Cref{fig:offline} presents the results for the small molecule design task; with additional experiments on proteins and material design in Appendix~\ref{app:offline_results}. 

\subsection{Online Data Acquisition}
\looseness=-1
In the online data acquisition setting, we additionally require an oracle (to score newly proposed candidates), and a search strategy (to propose new candidates). 
For the small molecule design task we use RDKit scorer \citep{rdkit} as oracle, and a single-edit SMILES search function that mutates the best candidate found so far. 
We demonstrate the same seven acquisition strategies run unchanged from the offline benchmark.

\begin{figure}[h]
\centering
\includegraphics[width=0.8\linewidth]{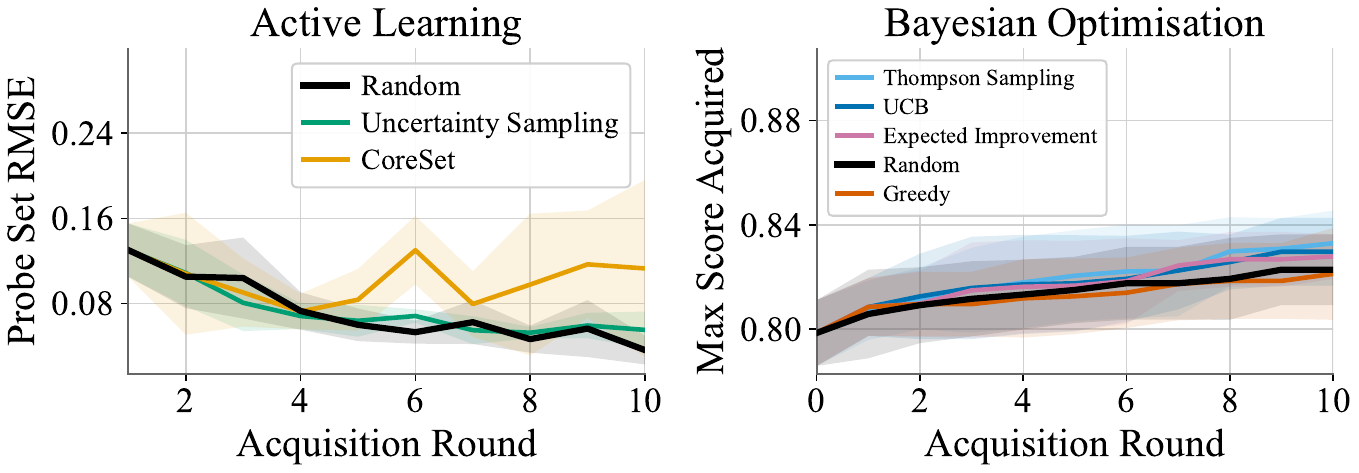}
\caption{\textbf{Small molecule design}. Online data acquisition for Active Learning objective (left) and Bayesian optimisation (right), against an \textsc{RDKit} oracle, with single-mutation search.
Results are the mean over $5$ random seeds; error bars are $\pm 1$ std.}
\label{fig:online}
\end{figure}

\Cref{fig:online} presents the results for the small molecule design task; with additional experiments on proteins and material design in Appendix~\ref{app:online_results}, which use \textsc{PyRosetta} \citep{chaudhury2010pyrosetta} and a \textsc{MACE} \citep{batatia2022mace} model as oracles, respectively. 

\subsection{Impact}
\label{subsec:further_case_studies}
\looseness=-1
We highlight \textsc{ALF}'s value to cutting edge research across scientific domains for both active learning and Bayesian optimisation objectives. 

\paragraph{Active learning for interatomic potentials.} 
Machine learning interatomic potentials (MLIP) can surrogate for expensive quantum chemistry calculations. 
In \cite{vargaumbrich2026pretrained, vargaumbrich2026forceaware}, authors used \textsc{ALF} to run offline active learning and compare data-acquisition strategies for reactive chemistry (Appendix~\ref{app:case_studies_results}, \cref{fig:casestudies}a). The approach demonstrates applying the neural tangent kernel as an acquisition signal and obtains lower energy and force error per round than prior state-of-the-art methods. The acquisition functions were added by extending the \texttt{AcquisitionFunction} base class, and utilising the surrogate, search, and design task classes.

\paragraph{Preference based protein sequence design.} In protein design, the aim is to find high-fitness sequences. In~\cite{hawkins2026likelihood}, authors used an early version of \textsc{ALF} for multi-round BO, extensively extending the Surrogate classes with pre-trained protein language models (PLM) ESM2, ProteinNPT~\citep{notin2023proteingym} and other baselines shown in (\cref{fig:casestudies}c). 
Their results show that fine-tuning a PLM with a ranking-based loss function directly on model likelihoods can retrieve higher scoring candidates per round than traditional MSE losses (Appendix~\ref{app:case_studies_results}, \cref{fig:casestudies}b). 

\section{Conclusion}
\label{sec:conclusion}
\looseness=-1

We present ALF, an ask/tell framework for scientific data acquisition and discovery. 
It utilises a single, clean API for \textit{offline} design tasks -- to validate methods against an existing known dataset, and \textit{online} acquisition campaigns against an oracle computing ground truth values on the fly. 
ALF is built from five easily extendible, data-agnostic components: datasets, surrogates, acquisition functions, search functions, and oracles. 
In this paper we highlight \textsc{ALF}'s current capabilities, comparing a range of acquisition strategies under both active learning and Bayesian optimisation objectives, across protein sequences, small molecules, and materials, and celebrated it's adoption in recent cutting edge research. 
Appendix~\ref{sec:limitations} discusses the limitations of the framework and the campaigns reported here, and Appendix~\ref{app:broader-impacts} the broader impacts of the work.
ALF is open source and available for the scientific community. We hope that it makes answering the question of "what data to acquire next?" easier across scientific domains. 
\section{Acknowledgments}

We thank Conor Finlay and Manus McAuliffe for their help during the development of the codebase; Akash Sinha for supporting the Bayesian optimisation pytorch implementations; Jack Simons, Zachary Weller-Davies, Jakob Kmec, Oliver Bent and Eszter Varga-Umbrich for using, reviewing and providing feedback on the codebase.

\newpage
\bibliographystyle{plainnat}
\bibliography{references}


\newpage
\appendix
\section{The ALF Library}
\label{app:library}

\subsection{Package structure}
\textsc{ALF} is deliberately split into two packages with different functionalities and dependencies. \texttt{alf-core} contains the different tasks (zeroshot, supervised and mult-round design), the abstract base classes for each component, the metrics, and the logging. It's dependencies only include lightweight libraries (such as, \texttt{numpy} and \texttt{pandas}) and pulls in no machine learning framework, so the interfaces and tasks can be imported, and new components written against them, without heavy ML stack dependencies. \texttt{alf-tools} maintains concrete implementations of each of the components introduced. It depends on \texttt{alf-core}, \texttt{torch}, \texttt{gpytorch}, and \texttt{botorch}, used by the GP surrogate and the acquisition backend. The domain-specific libraries are optional extras, installed only for the modality in use: \texttt{rdkit} for the molecule oracle, \texttt{matbench} and \texttt{pymatgen} for materials, \texttt{transformers} for the ESM-2 and ESMFold models, \texttt{chemprop} for the MPNN surrogate, and \texttt{mlip} for the MACE potential. Workflow umbrellas group these by domain, so \texttt{alf\_tools[protein]}, \texttt{[molecule]}, or \texttt{[materials]} install exactly the dependencies that modality needs and nothing more.

\subsection{Base Classes}
Each component has an abstract base class in \texttt{alf-core}. \Cref{tab:contracts} details each component and describes what functionality is provided by the base class and what each user needs to implement when extending the base class. Two of the components are not subclassed at all: the \texttt{Oracle} and the \texttt{Surrogate} are thin wrappers, configured by passing them a model.

\begin{table}[h]
\centering
\caption{What each component requires from a subclass, and what the base class
provides. \texttt{Oracle} and \texttt{Surrogate} are configured rather than
subclassed.}
\label{tab:contracts}
\small
\begin{tabular}{lll}
\toprule
Component & You implement & Provided by the base \\
\midrule
Dataset & \texttt{load\_dataset} & splitting, \texttt{query}, split updates, metrics \\
Model & \texttt{featurise}, \texttt{train}, & \texttt{setup}, embedding hook, metric \\
 & \texttt{predict}, \texttt{sample} & collection; reuse across roles \\
Acquisition fn. & \texttt{\_\_call\_\_(candidates, state)} & top-$k$ batching, optimizer integration \\
Search fn. & \texttt{\_\_call\_\_(state)} & dataset / generator / protocol subclasses \\
Oracle & --- (wrap a dataset or model) & offline/online branching \\
Surrogate & --- (wrap a model) & fit / predict / embed delegation \\
\bottomrule
\end{tabular}
\end{table}

A few dataclasses recur in these signatures: a \texttt{Candidate} is one data point, \texttt{LabelledCandidates} pairs candidates with their values, \texttt{Predictions} holds a surrogate's posterior distribution (means and optional variances), and \texttt{State} carries the datasets, surrogate, and acquisition history through the loop.

\paragraph{One model, three roles.}
The oracle, the surrogate, and generative search all take as input a \texttt{BaseModel} rather than reimplementing one. The oracle's \texttt{scorer} can be a \texttt{BaseModel} (scoring candidates on demand through \texttt{predict}); the surrogate wraps a \texttt{BaseModel} to fit and predict; and \texttt{GeneratorSearch} takes a \texttt{BaseModel} and proposes candidates through \texttt{sample}. A model therefore needs to be written once and can then act as the surrogate, the oracle, or the generator.

\subsection{ALF Tasks}
The ALF task drives the components through one run. The three task types; zeroshot, supervised and design share the same \texttt{BaseTask} scaffolding, the state, the evaluation, and the logging. \texttt{DesignTask} is the active learning loop, it runs the optimizer's ask/tell cycle for \texttt{num\_acq\_rounds}, the optimizer proposes a batch (\texttt{ask}), the oracle labels it, the surrogate is refit (\texttt{tell}), metrics are logged each round, and an aggregate summary is written at the end. \texttt{SupervisedTask} fits the surrogate once on a fixed train/validation split and evaluates on the test set. \texttt{ZeroShotTask} evaluates a pre-trained surrogate on the test set with no training at all.

\subsection{Offline to Online}
\label{app:offline_online}
The oracle and the search function are the only components that change between the two settings; the dataset, surrogate, optimizer, and task in \cref{fig:api} are untouched.
Offline, the search draws from the fixed pool and the oracle looks labels up in the dataset. Online, the search proposes de novo candidates and the oracle scores them with a model.
The code snippet below (\cref{lst:switch}) shows how to modify the search and oracle components to transfer an experiment from offline to online. 
 
\begin{lstlisting}[caption={Switching a campaign from offline to online: only the search function and the oracle change.},label={lst:switch}]
# Offline: draw from the pool, label by lookup
optimizer = Optimizer(acquisition_fn=UCB(), search_fn=DatasetSearch())
oracle    = Oracle(scorer=dataset)
 
# Online: propose new candidates, label with a model
optimizer = Optimizer(acquisition_fn=UCB(), search_fn=GeneratorSearch(generative_model))
oracle    = Oracle(scorer=scoring_model)
\end{lstlisting}

\subsection{Extending ALF}
A new implementation of a component is added by creating a child class of the component's base class. For example, \cref{lst:extending} below shows how a new acquisition function can be implemented. 

\begin{lstlisting}[caption={Extending the \texttt{AcqusitionFunction} base class to implement a custom acquisition function.},label={lst:extending}]
class MyAcquisition(AcquisitionFunction):
    def __call__(self, search_candidates, state):
        preds = state.surrogate.predict(search_candidates)
        scores = my_utility(preds.means, preds.variances)
        return LabelledCandidates(search_candidates, scores)
 \end{lstlisting}

\subsection{Tutorials and Documentation}\label{subsec:tutorials}
The repository ships runnable notebooks and how-to recipes.\footnote{\url{https://instadeepai.github.io/alf/}}
Offline and online design tutorials walk through a full campaign in each setting. Additionally, there are per-surrogate tutorials cover the GP~\citep{williams2006gaussian}, CNN, MLP, ensemble, ESM-2~\citep{lin2023esm2}, Chemprop~\citep{yang2019chemprop}, and the MLIP~\cite{brunken2026machinelearninginteratomicpotentials} model; a set of how-to recipes shows how to add a dataset, a model, an acquisition function, or a search function; and tutorials on the ProteinGym~\citep{notin2023proteingym} and GuacaMol~\citep{brown2019guacamol} datasets.
The hosted documentation includes the full API reference and an installation guide.
\section{Experimental Details}
\label{app:exp_details}

\paragraph{Surrogate and kernels.}
The surrogate is an exact-inference Gaussian process. On molecules and materials it uses a Mat\'ern-5/2 kernel with automatic relevance determination (ARD); on
sequences it uses a linear kernel, so the covariance is the $k$-mer spectrum inner product, with ARD off (per-dimension lengthscales would stop it being an inner product). The ARD arms place a dimension-scaled log-normal prior on the
lengthscales, $\mathrm{LogNormal}(\sqrt{2} + \log(d)/2,\ \sqrt{3})$, following
\citet{hvarfner2024vanilla}.

\paragraph{GP training parameters.}
Each round the GP is refit by maximising the marginal likelihood for 20 iterations with Adam at learning rate 0.5. Inputs are min-max normalised on molecules and materials and left unnormalised on sequences. Labels are $z$-scored on the training split for fitting and inverse-transformed before evaluation, so all reported RMSE is in original label units.

\paragraph{The active learning loop.}
Every campaign runs 10 acquisition rounds of 25 candidates, for 250 acquired per
cell, over five seeds. A seed fixes the data splits, the seed draw, and any stochastic acquisition.
The experiments are run on Nvidia H100 GPUs. 

\paragraph{Acquisition functions.}
Six acquisition strategies are used across the experiments with an additional random (uniform over the candidate set) baseline. The active learning style strategies are: CoreSet (greedy $k$-centres over the featurised inputs, label-blind), and uncertainty sampling ($\arg\max$ posterior standard deviation). The Bayesian optimisation strategies are: greedy ($\arg\max$ posterior mean), UCB ($\mu + \alpha\sigma$ with $\alpha = 1.0$), expected improvement, Thompson sampling (one posterior draw per selection).

\paragraph{Search functions.}
The online experiment has no fixed candidate pool, so candidates are generated each round by a modality-specific search function.
We use single mutant search for sequences, SMILES mutation search for molecules, and element substitution search for materials -- each the natural notion of a single-step edit in its domain: one amino-acid substitution, one character of a SMILES string, and one data-mined ionic species swap respectively.
In all three search strategies, the top-$K$ scoring candidates in the train dataset are taken as parents and mutated to form the candidate pool to acquire from. Note, any candidate in the pool that is already in the train or validation dataset is removed.
\section{Offline Benchmarking Results}
\label{app:offline_results}

\begin{figure}[h!]
\centering
\includegraphics[width=\linewidth]{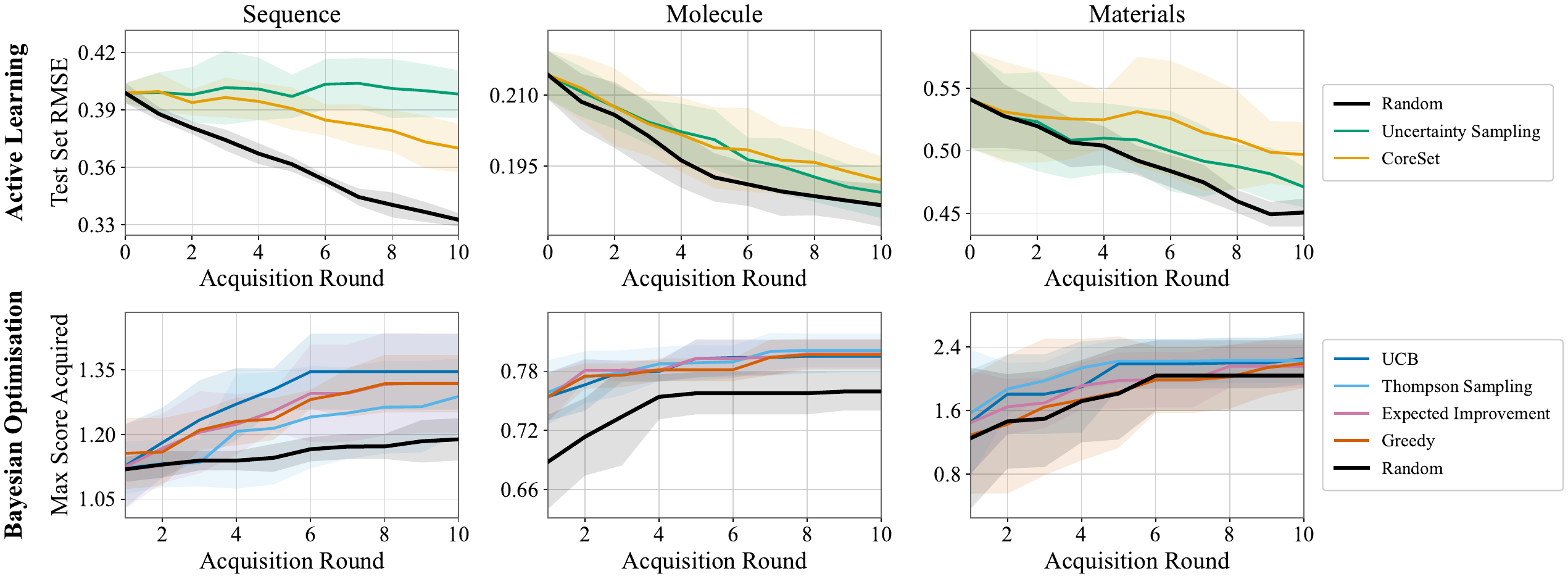}
\caption{Acquisition strategies across sequence, molecule, and materials under two objectives, from single \textsc{ALF} campaigns. \emph{Top, active learning:}
test-set RMSE against acquisition round (lower is better) for the information-based strategies. \emph{Bottom, Bayesian optimisation:} best score acquired so far (higher is better) for the utility-based strategies. Random is shown in both rows as the reference. Mean of five seeds, $\pm 1$ s.d.\ bands.}
\label{fig:offline_all_domains}
\end{figure}

\Cref{fig:offline_all_domains} compares the acquisition strategies across all three modalities
under both objectives. The pattern is consistent: under the active learning lens the information-based strategies stay close to random, which is strongest on all domains, and under the Bayesian optimisation lens the utility-based strategies acquire higher-scoring candidates than random on every modality.

\section{Online Data Acquisition Results}
\label{app:online_results}

\begin{figure}[h!]
\centering
\includegraphics[width=\linewidth]{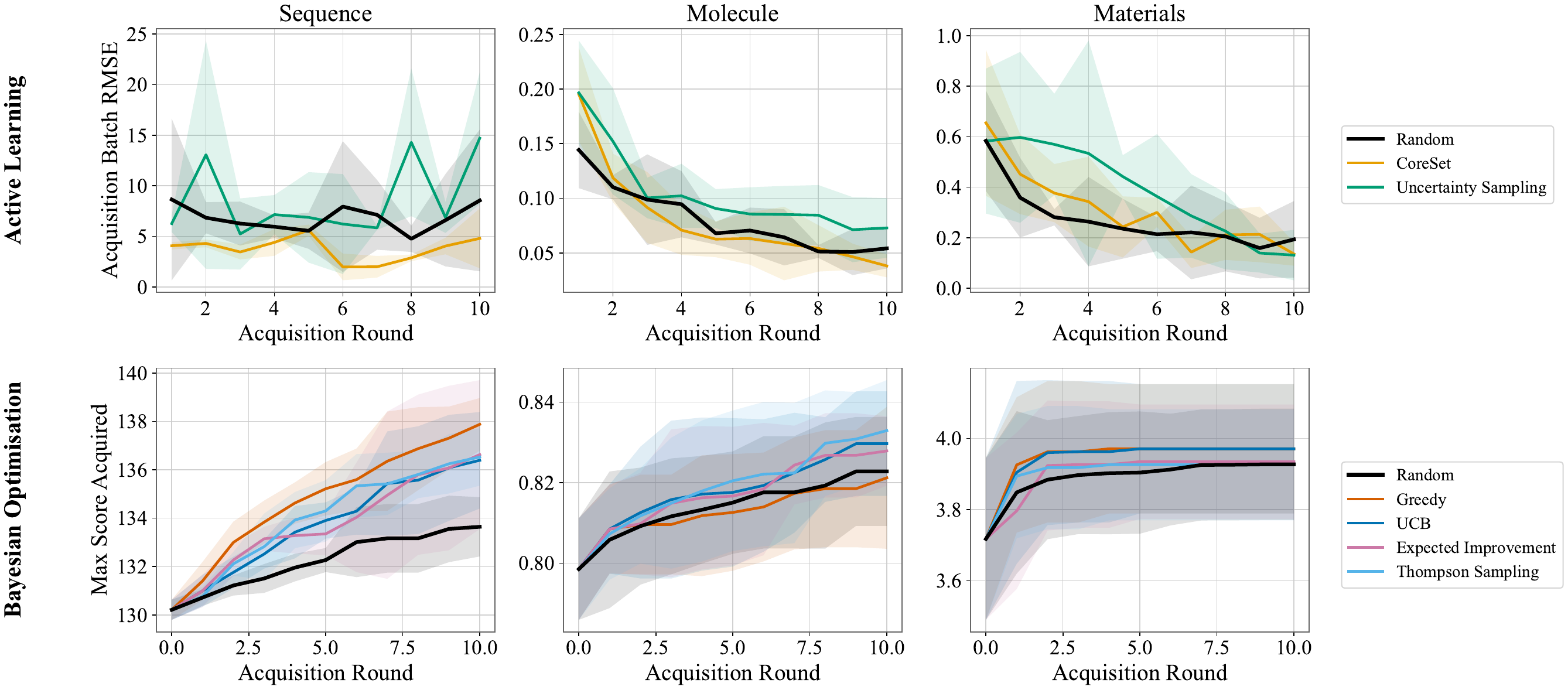}
\caption{Acquisition strategies across sequence, molecule, and materials under two objectives, from single online \textsc{ALF} campaigns. \emph{Top, active learning:}
to-be acquired batch RMSE against acquisition round (lower is better) for the information-based strategies. \emph{Bottom, Bayesian optimisation:} best score acquired so far (higher is better) for the utility-based strategies. Random is shown in both rows as the reference. Mean of five seeds, $\pm 1$ s.d.\ bands.}
\label{fig:online_all_domains}
\end{figure}

\Cref{fig:online_all_domains} repeats the online experiment across all three modalities, with the fixed pool replaced by a search that proposes candidates and an oracle model that scores them. Sequences use single mutant search against a PyRosetta~\citep{chaudhury2010pyrosetta} oracle, molecules use SMILES mutation search against an RDKit~\citep{rdkit} property scorer, and materials use element substitution search against a MACE~\citep{batatia2022mace} potential. Under the active learning objective, acquisition batch RMSE falls over the campaign for molecules and materials, but not for sequences. This may be attributed to the sequence surrogate and the search neighbourhood: single mutants share near-identical k-mer spectra, so the linear kernel separates little within a proposed batch. Under the Bayesian optimisation objective, every strategy improves the best scoring candidate found.

\section{Further Case Studies Results }
\label{app:case_studies_results}

\begin{figure}[h!]
\centering
\begin{subfigure}[b]{0.32\linewidth}
  \centering
  \includegraphics[width=\linewidth]{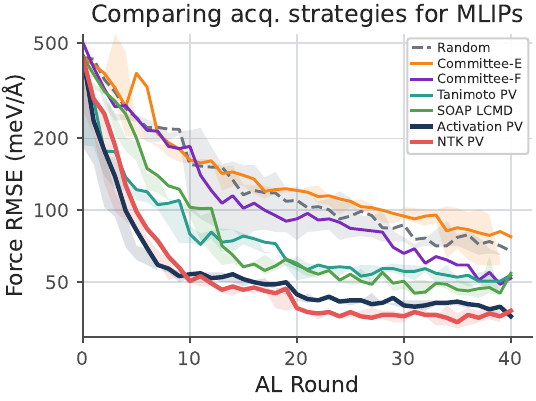}
  \caption{Acquisition strategies for MLIPs}
  \label{fig:case-mlip}
\end{subfigure}\hfill
\begin{subfigure}[b]{0.32\linewidth}
  \centering
  \includegraphics[width=\linewidth]{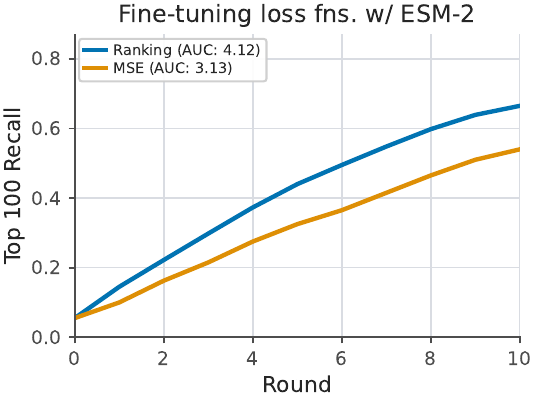}
  \caption{Ranking vs.\ MSE fine-tuning}
  \label{fig:case-ranking}
\end{subfigure}\hfill
\begin{subfigure}[b]{0.32\linewidth}
  \centering
  \includegraphics[width=\linewidth]{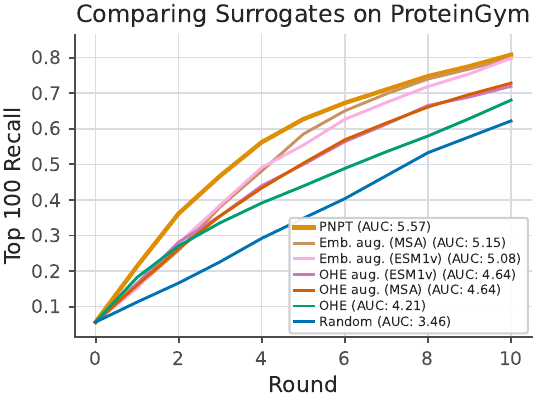}
  \caption{Comparing surrogate models}
  \label{fig:case-surrogate}
\end{subfigure}
\caption{\textsc{ALF} in published research. \textbf{(a)} Active learning for interatomic potentials~\cite{vargaumbrich2026pretrained,vargaumbrich2026forceaware}: force error per acquisition round, with a new neural-tangent-kernel signal among the strategies. \textbf{(b, c)} Bayesian optimisation for protein design~\citep{hawkins2026likelihood}: a ranking-based fine-tuning loss recovers more top candidates per round than MSE (b), and ProteinNPT \citep{notin2023proteingym} and other surrogates are compared in the same multi-round design setting (c).}
\label{fig:casestudies}
\end{figure}

\Cref{fig:casestudies} presents the results referenced in the case studies (\cref{subsec:further_case_studies}. Each panel is taken from the cited work and shows a multi-round campaign run through \textsc{ALF}, with performance plotted against the acquisition round.

Panel (a) shows force error on a reactive-chemistry dataset: the neural-tangent-kernel signal reaches a given error in fewer rounds than the committee, distance-based, and random baselines, and the gap widens over the campaign. Panels (b) and (c) show top-candidate recall on a protein design task. In (b), a ranking fine-tuning loss retrieves more of the top sequences than MSE at every round; in (c), the loop is fixed and the surrogate varied, with ProteinNPT~\citep{notin2023proteingym} and language model embedding baselines pulling ahead of one-hot encodings and random.

\section{Limitations}
\label{sec:limitations}

\textbf{Scope of the reported campaigns.} The experiments in \cref{sec:use-cases} are demonstrations of the framework rather than a study of acquisition strategies. Every campaign uses one exact-inference GP surrogate, one dataset per modality, and a single budget of 10 rounds of 25 candidates over five seeds. Five seeds is enough for error bands but not for ranking strategies whose curves overlap, and we make no claim that the orderings seen here hold at other budgets, batch sizes, or initial dataset sizes. Exact GP inference also scales cubically in the number of labels, so the surrogate used throughout is not the one we would recommend for campaigns much larger than those reported; ALF supports swapping it, but we do not benchmark that here.

\textbf{Oracles are computational stand-ins.} The online oracles are cheap models -- RDKit properties, PyRosetta energies, and a MACE potential -- standing in for the wet-lab assays and expensive simulations that motivate the work. They are fast, deterministic, and always return a value. Real oracles are noisy, are sometimes unavailable for a proposed candidate, and return labels after a delay that can span rounds. ALF's loop is synchronous and assumes every proposed batch comes back labelled, so measurement noise, attrition, and asynchronous or delayed feedback are not modelled.

\textbf{Search is restricted to single-step edits.} The three search functions mutate top-scoring parents by one amino-acid substitution, one SMILES character, or one ionic species swap. This keeps each round in the neighbourhood of the current best candidates, which compresses the differences between acquisition strategies and offers no mechanism for escaping a local optimum.
\texttt{GeneratorSearch} exists for proposing candidates from a generative model, but no campaign using it is reported. Relatedly, batches are formed by taking the top-$k$ scoring candidates, so nothing prevents a batch of near-duplicates.

\section{Broader Impacts}
\label{app:broader-impacts}

\textbf{Intended benefit.} Progress in the domains ALF targets is often bounded by measurement budget rather than modelling ability. Better tooling for active learning means fewer measurements are needed to reach a given result, so a fixed budget goes further in areas such as therapeutic design, enzyme engineering, and materials for energy storage. Since ALF provides a clean API where the the same code runs offline against a benchmark and online against an oracle, a method validated in a paper can be deployed without reimplementation, which reduces the gap between benchmark and deployment behaviour. Releasing the framework openly also lowers the engineering cost of running such campaigns.

\textbf{Dual use.} A data acquisition loop is agnostic to its objective, so the same machinery that optimises a useful molecular or protein property optimises a harmful one. ALF contributes orchestration rather than capability: the surrogates, oracles, datasets, and scoring functions it wraps are already publicly available, and defining a harmful objective does not require ALF.

\textbf{Misplaced confidence in an automated loop.} A campaign is only as good as its surrogate, and an efficient optimiser attached to a miscalibrated model will spend a real experimental budget quickly and in the wrong direction. This is a practical rather than a societal risk, but it is the one most likely to occur. ALF logs surrogate metrics every round precisely so that a campaign drifting away from its objective is visible while it is running rather than after the budget is spent, and we would encourage users to treat these as a stop condition rather than as post-hoc diagnostics.

\textbf{Compute.} The fine-tuning of models, running the oracles and performing search through protocols or generative models carries a non-trivial energy cost.
However, in the settings we target, the cost of a simulation or a wet-lab measurement tends to be more expensive than the training and inference of these models.

\end{document}